\documentclass[sigconf]{acmart}

\AtBeginDocument{%
  }

\acmDOI{}

\acmConference[LAK '25]{15th International Conference on Learning Analytics and Knowledgei}{March 03--07, 2025}{Dublin, Ireland}
\acmISBN{}

\acmSubmissionID{217}

\usepackage[most]{tcolorbox}
\usepackage{tabularx}

\newcolumntype{L}[1]{>{\raggedright\arraybackslash}p{#1}}

\newcommand{\kathrin}[1]{\textcolor{black}{#1}}

\begin{document}

\title[Can AI grade your essays?]{Can AI grade your essays? A comparative analysis of large language models and teacher ratings in multidimensional essay scoring}
\author{Kathrin Seßler}
\email{kathrin.sessler@tum.de}
\orcid{0000-0002-3380-4641}
\affiliation{%
  \institution{Technical University Munich}
  \city{Munich}
  \country{Germany}
}

\author{Maurice Fürstenberg}
\email{maurice.fuerstenberg@uni-tuebingen.de}
\affiliation{%
  \institution{University of Tübingen}
  \city{Tübingen}
  \country{Germany}}

\author{Babette Bühler}
\affiliation{%
  \institution{Technical University Munich}
  \city{Munich}
  \country{Germany}
}

\author{Enkelejda Kasneci}
\affiliation{%
  \institution{Technical University Munich}
  \city{Munich}
  \country{Germany}
}



\begin{abstract}
The manual assessment and grading of student writing is a time-consuming yet critical task for teachers. Recent developments in generative AI, such as large language models, offer potential solutions to facilitate essay-scoring tasks for teachers. 
In our study, we evaluate the performance and reliability of both open-source and closed-source LLMs in assessing German student essays, comparing their evaluations to those of 37 teachers across 10 pre-defined criteria (i.e., plot logic, expression). 
A corpus of 20 real-world essays from Year 7 and 8 students was analyzed using five LLMs: GPT-3.5, GPT-4, o1, LLaMA 3-70B, and Mixtral 8x7B, aiming to provide in-depth insights into LLMs' scoring capabilities. 
Closed-source GPT models outperform open-source models in both internal consistency and alignment with human ratings, particularly excelling in language-related criteria.
The novel o1 model outperforms all other LLMs, achieving Spearman’s $r = .74$ with human assessments in the overall score, and an internal consistency of $ICC=.80$.
These findings indicate that LLM-based assessment can be a useful tool to reduce teacher workload by supporting the evaluation of essays, especially with regard to language-related criteria. 
However, due to their tendency for higher scores, the models require further refinement to better capture aspects of content quality.
\end{abstract}



\begin{teaserfigure}
\centering
  \includegraphics[width=\textwidth]{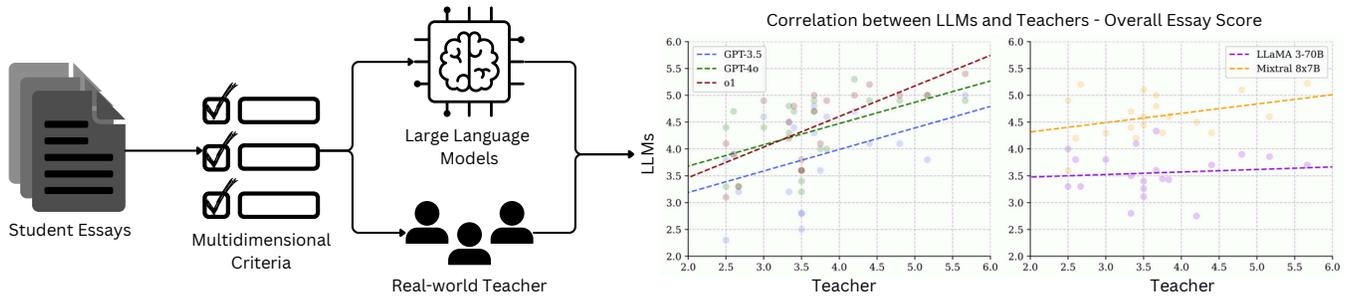}
  \caption{Design and workflow of our study: student essays are evaluated based on predefined criteria by both Large Language Models and human teachers, followed by a comprehensive analysis of the resulting ratings.}
  \Description{Design and workflow of our study}
  \label{fig:teaser}
\end{teaserfigure}


\maketitle

\section{Introduction}
The correction and evaluation of student texts is a tedious yet crucial process for teachers, especially in subjects where a significant amount of assessment-relevant text is produced. One of the relatively few studies of teachers' working hours in German schools revealed, for example, that correcting students' texts is enormously time-consuming \cite{arbeitszeitstudie_2016}. 
The study found that, after teaching itself (29\%) and the preparation and follow-up of lessons (22\%), proofreading (including feedback to content etc.) accounted for the third-highest proportion of working time for secondary school teachers, at 14\%. Notably, the study did not differentiate between subjects, hence teachers of less correction-intensive subjects were also included in this proportion. 
For example, a typical German teacher with 25 students, three German classes, and four class tests per year — plus one practice essay per test — would need to correct around 600 essays annually, averaging more than two per working day — and that's just for the subject German.

Despite the time and effort teachers invest in correcting student texts, the feedback provided is often delayed, and extremely heterogeneous \cite{muller2023qualitat}. Currently, Germany, like many other countries, is also struggling with a systematic shortage of teachers \cite{KMK2023}, which is expected to worsen in the future. In this context, supporting time-consuming tasks like essay scoring is crucial, and technical solutions can offer valuable assistance to alleviate the workload on teachers.

The rapid advancements of Artificial Intelligence (AI) and Large Language Models (LLMs), such as GPT-4 and o1 by OpenAI~\cite{GPT4,o1}, LLaMA 3 by Meta \cite{llama3modelcard}, or Mistral AI \cite{jiang2024mixtral} open various new possibilities, also within the educational context \cite{kasneci2023chatgpt}. These models possess the capability to process, analyze, and generate natural language with high proficiency, offering the potential to enhance teaching and learning processes. For instance, LLMs can support lesson preparation through automated question generation \cite{bhat2022towards}, facilitate teacher collaboration via conversational AI \cite{ji2023systematic}, and support the evaluation and feedback generation for student errors \cite{bewersdorff2023assessing}. Moreover, the abilities of LLMs extend across various educational domains, including language learning \cite{munoz2023examining}, mathematics \cite{nguyen2023evaluating}, and life science education \cite{bewersdorff2024taking}.
By integrating these advanced AI tools, educators can devote more time to other important tasks, enhance the quality and consistency of student assessments, and foster more interactive and personalized learning experiences.

The challenges mentioned above, coupled with ongoing advancements in AI, raise the question of whether LLMs can serve as efficient alternatives or at least provide valuable support in the evaluation of student essays. 
Although AI-generated feedback is progressively being integrated into educational applications, there remain substantial gaps in our understanding of the quality and effectiveness of this feedback.
While some initial approaches employ AI to provide feedback on student texts \cite{sessler2023peer, fiete2024} and others focus on Automated Essay Scoring (AES) \cite{ramesh2022automated,pado2023}, a \textbf{detailed, criteria-based evaluation} is still missing.
Previous research, mostly evaluating texts based on a holistic score \cite{sawatzki2021deep,pado2023} or concentrating on few general and task-agnostic criteria \cite{mizumoto2023exploring,naismith2023automated}, often falls short of fully capturing the complexity and details inherent in student writing, particularly within the context of German language lessons.
Also, the limited availability of suitable data resources and the \textbf{uncertainty around a clear and consistent ground truth}—typically based on teacher evaluations, which are often subjective and require multiple raters—arise as further challenges. For instance, the widely used ASAP dataset~\cite{asap_dataset} relies on only two raters for most of the essays. While this setup is standard for many datasets, it could limit the robustness of the annotations when evaluating highly subjective dimensions like essay quality.
This uncertainty also extends to the ratings of automated essay scoring system. Only few studies have examined the \textbf{consistency of LLMs} across multiple runs when evaluating a text \cite{hackl2023gpt}.
Finally, existing research lacks a \textbf{qualitative analysis of the reasoning processes} of LLMs to improve the understanding of which essays aspects have an influence on the model assessment.

To leverage LLMs to their full potential in educational settings, we address these gaps by evaluating the quality of open- and closed-source LLMs in detail by assessing student text based on ten predefined language- and content-related criteria. Therefore, we conducted a user study involving 37 German teachers from different schools and educational levels, collecting on average 5.45 ($SD=0.92$) ratings per 20 German real-world student essays. 
Specifically, we compare the human ratings with those generated by GPT-3.5, GPT-4 \cite{GPT4}, o1 \cite{o1}, LLaMA 3-70B \cite{llama3modelcard} and Mixtral 8x7B \cite{jiang2024mixtral}, thoroughly analyzing their strengths and limitations across different evaluation categories and their reasoning process through inter-category correlation examination. We underpin our results by including LLM reliability measurements through comparing repeated assessments.
Our analysis adopts a multidimensional approach, extending beyond the mere holistic score to examine biases and performance variations in different assessment aspects and gain in-depth insights into the capabilities of the tested models. 

This comprehensive evaluation aims to understand how well LLMs align with human evaluations and to identify areas where they excel or require improvement.
To achieve these goals, we target the following research questions:

\begin{itemize}
    \item \textbf{RQ1:} How reliably do open-source and closed-source LLMs perform in essay evaluation, and how do their assessments correlate with real-world teacher evaluations of German-language texts?
    \item \textbf{RQ2:} What are the strengths and limitations of using LLMs to assess multidimensional aspects of essay quality beyond providing a basic holistic score?
    \item \textbf{RQ3:} How do different evaluation criteria influence the reasoning process for the overall essay scores of humans and LLMs?
\end{itemize}


\section{Related Work}

\subsection{Text assessment}
When evaluating student texts, a scale can be established with 'holistic' at one end and 'analytical' at the other \cite{schroter2022}. This distinction relates to the nature of the judgment process: holistic approaches typically rely on implicit criteria that are difficult to articulate, while analytical methods are guided by explicit criteria, often structured in detailed rubrics. Despite the scarcity of studies—particularly regarding the assessment of German school texts—research suggests no significant differences in objectivity between these two approaches \cite{grzesik1984, schroter2022}. Both holistic and analytical assessments are subject to variability between different evaluators \cite{birkel2002, schroter2022}, though there is also substantial agreement between the judgments derived from each approach \cite{pohlmann2016}. Analytical assessments tend to be more rigorous and less prone to judgment errors \cite{schroter2022}. Note, however, that the study situation for German texts is very sparse and that the various studies were carried out in completely different age groups and text types in different years and are therefore not comparable. 
From the learners' perspective, criteria-based feedback is key to improving their writing skills, therefore, in this study, we focus on analytical feedback on student texts.


\subsection{LLMs for text assessment}
\paragraph{Traditional Automated Essay Scoring}
Automated Essay Scoring (AES) techniques have been developed since 1966 \cite{ramesh2022automated}. Primarily, they exploited statistical features to analyze the text \cite{ke2019automated}. Then, with the introduction of deep learning methods, it became possible to concentrate on more advanced syntactic and semantic features and to analyze the style and content of a text \cite{bai2023survey}, applying LSTMs or Transformer-based models like BERT \cite{devlin-etal-2019-bert}. 
For example, BERT has been employed to extract features to train a regression model \cite{doewes2021limitations}, directly output a class label \cite{sung2019improving,xue2021hierarchical}, or RoBERTa and Bi-LSTM have been combined to assess essays \cite{beseiso2021novel}. Further, hand-crafted features in combination with existing LSTM or BERT structures have been shown to boost the performance \cite{uto2020neural}.
But overall, the models up to 2022 concentrate more on language than on content and do not emphasize cohesion and coherence of the essays enough \cite{ramesh2022automated}. Also, many works rely on statistical features to determine one holistic final score and concentrate on English data.

\paragraph{Short Answer Grading on German data}
To adapt for German essays, \cite{sawatzki2021deep} applied BERT, focusing on Automatic Short Answer Grading (SAG) (with less than 100 words on average). They achieved a Pearson correlation coefficient of .75, indicating a strong alignment between the model's predictions and human ratings. Similarly, \cite{pado2023} examined the inter-rater reliability between SBERT and human ratings for German corpora, also concentrating on SAG. Although they simplified the task by reducing the grading to a binary decision and training the model on a specific dataset, the transformer model yielded an average precision of only 71.4\%. Unlike free-text essay scoring, however, the evaluation of short answers typically relied on a predefined ground truth for each question. This distinction raised concerns about the suitability of transformer models for more complex tasks such as the assessment of student texts using expert ratings.
\kathrin{There is an absence of studies focusing on the assessment of standard German essays that go beyond holistic scoring.}

\paragraph{Applying GPT models on English Essay data}
The rise of LLMs in 2022, has opened up new opportunities for automatic essay grading. One of the first studies was conducted by \cite{chiang-lee-2023}, which compared the ratings of three teachers on 400 English text fragments (averaging 150 words) with the ratings of GPT-3 \cite{ouyang2022training}. The three teachers and the model rated the texts – half written by humans and half generated by GPT-2 \cite{radford2019language} – on a five-point Likert scale in four categories: grammaticality, cohesion, sympathy, and relevance. 
The ratings of the teachers and the model only correlated strongly for the criterion relevance, 
while the other criteria showed a weak positive correlation.
\cite{mizumoto2023exploring} leveraged GPT-3 to evaluate over 12,000 essays from the TOEFL11 dataset \cite{blanchard2013toefl11} across four dimensions: task response, coherence and cohesion, lexical resources, and grammatical range. They compared GPT-3’s performance with 45 linguistic features and discovered that while the linguistic features alone outperformed GPT-3, the combination of both approaches yielded the best results. Specifically, GPT-3 achieved an accuracy of 54\% compared to professional human evaluators but demonstrated an 89\% agreement within a deviation of 1 to 2 points.
\cite{floden2024grading} reported similar findings with GPT-3.5, showing perfect agreement with human raters on only 30\% of exams. However, GPT-3.5 maintained a 70\% accuracy within a 10\% range around the true score, indicating reasonable but not perfect alignment with human judgments.

In the context of discourse coherence analysis, \cite{naismith2023automated} identified a 56\% exact match and a 97\% adjacent match between human evaluators and GPT-4 \cite{GPT4}. Additionally, both \cite{alers2024using} and \cite{pinto2023large} employed GPT-4 for open-ended question grading. \cite{alers2024using} found a strong correlation between human and LLM scores, while \cite{pinto2023large} demonstrated that the model’s feedback closely resembled expert feedback. \cite{lee2024applying} compared GPT-3.5 and GPT-4 for automated scoring in science education, revealing that prompting strategies such as few-shot learning and chain-of-thought (CoT)  \cite{wei2022chain} can enhance accuracy.

All these studies rely on closed-source GPT models and focus exclusively on English text data. Moreover, many provide only holistic scores or assess a limited set of artificial criteria. There is a notable lack of detailed analyses exploring how LLM-generated scores differ from human judgments across multidimensional criteria. 

\paragraph{Applying open-source LLMs on Essay data}
\cite{stahl2024} investigated the use of open-source LLMs for generating feedback. They evaluated different prompting strategies for zero- or few-attempt learning to determine how well Mistral 7B \cite{jiang2023mistral} could generate essay feedback. This approach of comparing different prompting strategies appeared promising for future research. The study showed that combining automated essay scoring (AES) with feedback generation could improve scoring performance, although the overall impact of AES on feedback quality was minimal. 
Both, LLM-based scores and manual scores, were used to evaluate the usefulness of the feedback.
However, using another LLM to evaluate a feedback-generating LLM raises concerns, such as the risk of perpetuating the biases inherent in the models and the lack of the nuanced understanding of a human expert, such as a teacher. Furthermore, the study did not report the qualifications or backgrounds of the 12 human raters, which is a serious omission. Understanding these raters' expertise is essential to assessing the quality of manual feedback rating. It raises the question of how amateur raters might differ from an LLM in terms of feedback quality.

Acknowledging the limited research in open-source LLMs for automated essay scoring, the scarcity of non-English datasets, and the absence of multidimensional, didactic-based evaluation criteria, our study aims to conduct a comprehensive evaluation of various closed- and open-source LLMs in multidimensional essay assessment and gather new insights into their reasoning processes.
In addition, almost no study uses authentic teacher ratings in combination with real learner texts, which impairs the ecological validity of previous studies.


\section{Methodology}

In our study, we aim to analyze the performance of LLMs in evaluating student texts according to ten pre-defined criteria. In the following, we describe the details of the study, including the essays and scoring criteria used, the participants, the application of the LLMs, and the metrics employed for analysis.

\begin{table*}[htbp]
\centering
\begin{tabularx}{\textwidth}{lL{3cm}cX}
\toprule
No. & Title & Type & Description \\
\midrule

1 & Headline & C & The headline creates suspense and is appropriately chosen. \\
2 & Introduction & C & The introduction is interesting and builds up the suspense.\\
3 & Main part  & C & The main part builds up the suspense convincingly and skillfully and contains a turning point.\\ 
4 & Conclusion & C & The conclusion rounds off the story and brings all the narrative strands to a close. \\
5 & Verbal images \& descriptive elements & L & The choice of words is vivid and linguistic images / descriptive elements are used.\\
6 & Literal speech \& inner monologue & L &Literal speech and inner monologue are used purposefully.\\
7 & Plot logic  & C & The plot is structured without logical breaks.  \\
8 & Expression \& sentence structure & L &The expression and sentence structure are varied and convincing. \\
9 & Spelling \& punctuation & L & The spelling and punctuation are secure.\\
0 & Overall judgment & C/L & Overall, the work is... \\
\bottomrule
\end{tabularx}
\caption{\label{tab:categories_description}The 10 evaluation criteria, categorized into content-related (C) and language-related (L) aspects, each assessed using a six-point Likert scale.}
\end{table*}

\subsection{Student Essay Dataset}

The text corpus comprises $N=20$ real-world student texts from pupils in Year 7 ($n=10$) and Year 8 ($n=10$) at two secondary schools (one ``Gymnasium'' and one ``Realschule'') in Germany. These essays were written as part of a performance assessment. Pupils in the seventh grade wrote a narrative with descriptive elements based on the ballad "The Sorcerer's Apprentice" (in German ``Der Zauberlehrling'') by J. W. Goethe. Those in the eighth grade wrote narratives inspired by two paintings by Edward Hopper (``Nighthawks'' and ``Gas''). 
%
To ensure representativeness despite a small corpus, we selected narratives since they are taught in all German school types.

\subsection{Teacher Essay Scoring}
The 20 student essays were presented to $N=37$ teachers with the task of reading the texts and assessing them according to 10 pre-defined criteria using a six-point Likert scale, ranging from \textit{Not true at all} to \textit{Fully applies}. The six-point Likert scale was selected to align with the German grading system, which also employs six levels, enabling teachers to provide consistent and familiar ratings. In addition, the even number of possible choices eliminates the risk of a tendency towards the center \cite{vanHerk2004}. 

Since no universally standardized criteria exist for essay evaluation for any type of text in German, the criteria for this study were derived from real-world teacher feedback specific to these essay types and were later refined and formalized by an expert in German didactic. The advantage of this approach, apart from the increase in ecological validity, is that teacher are familiar with the categories, which in turn has a positive influence on the reliability of their ratings. Half of the resulting categories asses the content (i.e., \textit{Introduction}, \textit{Main Part}, and \textit{Conclusion}), while the others focus on the writing style and language (e.g., \textit{Verbal images} and \textit{Descriptive elements}, \textit{Literal speech \& inner monologue} and \textit{Spelling \& punctuation}), with one category assessing overall judgment, as detailed in Table \ref{tab:categories_description}.
The teachers submitted their assessments anonymously via an online portal. Each teacher was asked to rate three texts; however, there are some missing values, resulting in a total of 1,090 human ratings across the 20 texts and the 10 assessment categories. Each essay received between three and seven ratings from different teachers, with an average of $5.45 \pm 0.92$ ratings per essay. 
We employ the average of the teacher ratings as our ground truth per essay and evaluation criteria.



\subsection{LLM Essay Scoring}
To automatically evaluate student essays based on the ten pre-defined criteria, we selected different LLMs, to compare the performance of various foundation models.
For closed-source models, we used GPT-3.5 (\texttt{gpt-3.5-turbo-0125}), GPT-4 (\texttt{gpt-4o-2024-05-13}) \cite{GPT4} and o1 (\texttt{o1-preview}) \cite{o1} as representative examples. These GPT models were integrated into our evaluation pipeline via the OpenAI API.
For open-source LLMs, we chose LLaMA 3-70B \cite{llama3modelcard} and Mixtral 8x7B \cite{jiang2024mixtral}. 
Initial experiments included smaller variants of these models; however, their performance was inferior to their larger counterparts. Therefore, the smaller models were excluded from further analysis. Both open-source models were executed locally in half-precision using two NVIDIA A100 80GB GPUs to ensure efficient processing.

All LLMs were prompted using a zero-shot approach, instructing them to evaluate one text at a time. For each essay, a new conversation was initiated to maintain the independence of evaluations. The prompt structure was designed to align with the predefined evaluation criteria, ensuring consistency across all model assessments. This methodology allowed us to systematically compare the performance of different LLMs in a controlled and unbiased manner.

\begin{figure}[htbp]
    \centering
\begin{tcolorbox}[colframe=black!70!white, colback=black!5!white]
\small{
\texttt{You are a teacher. Analyze the essay written by a 13-year-old child according to the given criteria. Return a scalar number from 1 to 6 for each criteria. 1 means the criteria is not fulfilled at all, 6 means it is completely accomplished. Return only a JSON. \#\# Criteria = \{criteria\}; \#\# Essay = \{text\}; \#\# Rating = }
}
\end{tcolorbox}
\caption{\label{fig:prompt}Standardized zero-shot prompt employed across all LLMs to ensure a fair comparison of their essay evaluation performance.}
\end{figure}

\noindent
The prompt, illustrated in Figure \ref{fig:prompt}, specifies the model's role as a teacher, provides the context of an essay written by a 13-year-old student, outlines the specific task of analyzing the essay according to predefined criteria, and defines the desired output format as JSON. This prompt was consistently used throughout all experiments to ensure uniformity. While variations in input prompts can lead to different results \cite{stahl2024}, specific prompt engineering was not the focus here.

To analyze the reliability of the predictions and capture the stochastic nature of the models, each text was evaluated ten times by each LLM, using a temperature of 0.7.
The mean of these ten assessments represents the average rating assigned by the LLMs, analogous to the average of multiple human ratings for the same text. 
For open-source models, we filtered out any missing values or predictions that fell outside the predefined range ($N=169$). In contrast, the GPT models consistently returned outputs in a valid format, ensuring the integrity of the collected data. A total of $N=9,931$ LLM ratings were collected.

\subsection{Analysis}
To investigate our research questions, we systematically analyzed the evaluation outputs generated by both human raters and LLMs.
\textbf{RQ1} addresses the reliability and quality of the evaluations. To answer this, we compared the different runs of each model on the exact text by calculating inter-class correlation, treating each run as an individual rater (see Section \ref{sec:irr}). For the following sections, the paper always considers the average of the ten runs as the final rating. In Section \ref{sec:correlation_analysis}, we compared the assessments of the LLMs to those of real-world teachers using the Spearman correlation coefficient to identify similarities between the ratings.
For \textbf{RQ2}, we examined the overall holistic scores alongside all multidimensional aspects, including both language- and content-related criteria (see Section \ref{sec:overall_evaluation}). By comparing the distribution of ratings between LLMs and human raters and applying the Mann-Whitney U test, we identified any differences in how specific criteria are rated by each group.
Lastly, \textbf{RQ3} explores the factors influencing the reasoning processes behind the evaluations. In Section \ref{sec:intra_criteria}, we analyzed inter-category correlations for both human raters and LLMs. This analysis helps us understand which categories have the strongest associations with the overall score, providing insights into the reasoning processes of both humans and LLMs and identifying which criteria most significantly influence the holistic assessment.


\section{Results}

In the following section, we explore the discrepancies between real-world teacher assessments and open- as well as closed-source LLM-generated evaluations of student texts, aiming to address our three key research questions.


\subsection{Reliability of model predictions} \label{sec:irr}
We compare multiple runs of the same prompt to assess the reliability of each model's evaluations in addressing RQ1. Consequently, we obtain ten ratings for each data point, similar to human raters. Table \ref{tab:icc} presents the ICC statistics for all employed foundation models.

\begin{table}[htbp]
    \centering
    \begin{tabular}{cccccc}
    \toprule
          & GPT-3.5 & GPT-4 & GPT-o1 & LLaMA & Mixtral  \\ 
         \midrule
         ICC & 0.84 & 0.73 & 0.80 & -0.04 & 0.01 \\ 
    \bottomrule
    \end{tabular}
    \caption{ICC values comparing the inter-rater reliability of each LLM across multiple evaluation runs.}
    \label{tab:icc}
\end{table}


It is evident that the closed-source models demonstrate a fair level of consistency in their ratings across individual runs, as indicated by 
ICC showing moderate to good reliability (.73 - .84) \cite{koo2016guideline}.
In contrast, the open-source models LLaMA 3 and Mixtral, display significant variability, with low values and poor agreement between multiple runs. This inconsistency needs to be taken into account in the subsequent analysis.


\subsection{Correlation Analysis Between LLM Evaluations and Teacher Assessments}\label{sec:correlation_analysis}


Continuing our examination of RQ1, we compare the assessment of LLMs and humans by computing the Spearman correlation coefficients $r$ for all evaluation criteria in Table \ref{tab:pearson_correlation_results}.
o1 demonstrates the strongest correlation with human ratings, achieving significance in nine out of ten categories. Specifically, it is the only model demonstrating a high, significant correlation of .74 to teacher ratings in the overall category. GPT-4 maintains significant correlation in seven out of ten categories, with a moderate agreement with human raters in most of the categories, while GPT-3.5 reaches significant correlations in four categories. This again shows the improvement across subsequent model versions. In contrast, Mixtral shows only weak or non-significant correlations across all criteria. LLaMA 3 exhibits nearly zero correlation to teacher ratings and, in some cases, even negative correlations, underscoring its inconsistency with human evaluations.

\begin{table*}[htbp]
    \centering
\begin{tabular}{l|cc|cc|cc|cc|cc} 
\toprule 
 Category & \multicolumn{2}{c}{GPT-3.5} & \multicolumn{2}{c}{GPT-4} & \multicolumn{2}{c}{GPT-o1} & \multicolumn{2}{c}{LLaMA 3-70B} & \multicolumn{2}{c}{Mixtral 8x7B}\\ 
 & $r$ & $p$ & $r$ & $p$ & $r$ & $p$ & $r$ & $p$ & $r$ & $p$ \\ 
\midrule 
Overall & 0.418 & 0.067 & \textbf{0.575} & \textbf{0.008} & \textbf{0.742} & \textbf{0.000} & 0.091 & 0.703 & 0.311 & 0.182 \\ 
Headline & -0.005 & 0.984 & 0.159 & 0.504 & \textbf{0.699} & \textbf{0.001} & 0.131 & 0.581 & 0.252 & 0.284 \\ 
Introduction & 0.313 & 0.179 & 0.325 & 0.162 & 0.127 & 0.594 & 0.014 & 0.953 & 0.348 & 0.133 \\ 
Main part & 0.279 & 0.234 & 0.386 & 0.092 & \textbf{0.466} & \textbf{0.038} & -0.094 & 0.694 & 0.376 & 0.103 \\ 
Conclusion & \textbf{0.450} & \textbf{0.046} & \textbf{0.453} & \textbf{0.045} & \textbf{0.714} & \textbf{0.000} & -0.288 & 0.218 & 0.011 & 0.964 \\ 
Verbal image & \textbf{0.483} & \textbf{0.031} & \textbf{0.680} & \textbf{0.001} & \textbf{0.738} & \textbf{0.000} & -0.064 & 0.789 & 0.146 & 0.538 \\ 
Literal Speech & 0.138 & 0.561 & \textbf{0.521} & \textbf{0.019} & \textbf{0.805} & \textbf{0.000} & -0.262 & 0.264 & 0.375 & 0.103 \\ 
Plot logic & 0.425 & 0.062 & \textbf{0.585} & \textbf{0.007} & \textbf{0.608} & \textbf{0.004} & -0.032 & 0.893 & 0.442 & 0.051 \\ 
Expression & \textbf{0.520} & \textbf{0.019} & \textbf{0.626} & \textbf{0.003} & \textbf{0.675} & \textbf{0.001} & 0.177 & 0.455 & 0.211 & 0.372 \\ 
Spelling & \textbf{0.728} & \textbf{0.000} & \textbf{0.846} & \textbf{0.000} & \textbf{0.814} & \textbf{0.000} & 0.406 & 0.076 & 0.005 & 0.984 \\ 
\bottomrule 
\end{tabular} 

    \caption{\label{tab:pearson_correlation_results}Spearman correlation coefficients ($r$) and the corresponding $p$-values comparing human ratings with those of the five LLM-models. Significant correlations are highlighted in \textbf{bold}.}
\end{table*}

\begin{figure*}[htbp]
    \centering
    \includegraphics[width=\linewidth]{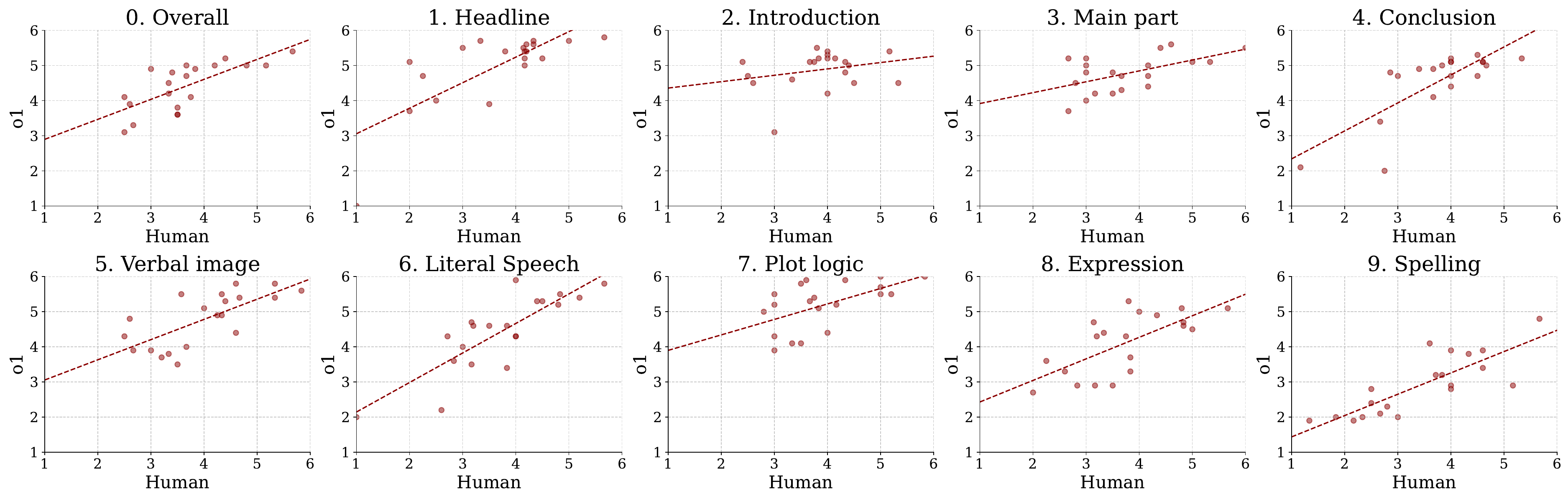}
    \caption{\label{fig:correlation_gpt4}Correlation between o1 and human ratings across all evaluation criteria. The red line represents the linear least-squares regression between the two sets of ratings.}
\end{figure*}

Figure \ref{fig:correlation_gpt4} visually shows the strong correlation between o1 and human ratings, especially for language-related features such as spelling ($r$=0.814), use of literal speech ($r$=.805), and verbal imagery ($r$=.738). Additionally, the overall rating exhibits a robust correlation despite having an intercept of 2.3. However, two content-related categories — \textit{Introduction}, and \textit{Main Part} — demonstrate only weak or moderate correlations. In these areas, o1 is less effective at distinguishing between higher and lower-quality essays. 

\subsection{Rating Comparison between LLM and Human Evaluation}\label{sec:overall_evaluation}

To address RQ2, we first compare the distribution of the ratings in the \textit{Overall} category made by teachers and the LLMs, as shown in Table \ref{tab:overall_results}. The human raters exhibit a lower average rating than the GPT models, indicating a stricter assessment approach. The LLaMA 3 model presents a similar average to the human raters. However, the variance in ratings across essays is significantly higher for human raters than for all LLMs, with the LLaMA 3 and Mixtral displaying particularly low variance. This suggests that the open-source models assess essays more uniformly, clustering ratings around the midpoint, whereas human raters and the closed-source models provide a wider differentiation between essay qualities. Additionally, GPT-4, o1, and Mixtral provide significantly higher ratings than humans. 
The low rater correlation of both LLaMA 3 and Mixtral (Section \ref{sec:irr}) contributes to these observations by resulting in more average values and reduced variance.
This highlights a limitation in the consistency and alignment of certain LLMs with human evaluative standards. 



\begin{table*}[htbp]
    \centering
    \resizebox{\textwidth}{!}{
    \begin{tabular}{lc|cc|cc|cc|cc|cc} 
        \toprule
         & Human & GPT-3.5 & $p$ & GPT-4 & $p$  & GPT-o1 & $p$  & LLaMA 3 & $p$ & Mixtral & $p$\\
         \midrule
         Overall & 3.65 $\pm$ 0.86 & 3.85 $\pm$ 0.81 & 0.30 & \textbf{4.34 $\pm$ 0.66}  & \textbf{0.02} & \textbf{4.41 $\pm$ 0.68} & \textbf{0.01} & 3.55 $\pm$ 0.41 & 0.97 & \textbf{4.60 $\pm$ 0.41} & \textbf{0.00}\\
         \bottomrule
    \end{tabular}
    }
    \caption{\label{tab:overall_results}Overall average evaluations of all essays, including $p$-values from the Mann-Whitney U test comparing the distribution of the LLM scores to human scores. Significant differences are indicated in \textbf{bold}.}
\end{table*}


\subsection{Analysis of Rating Distributions Across Evaluation Categories}

\begin{figure*}[htbp]
    \centering
    \includegraphics[width=0.8\linewidth]{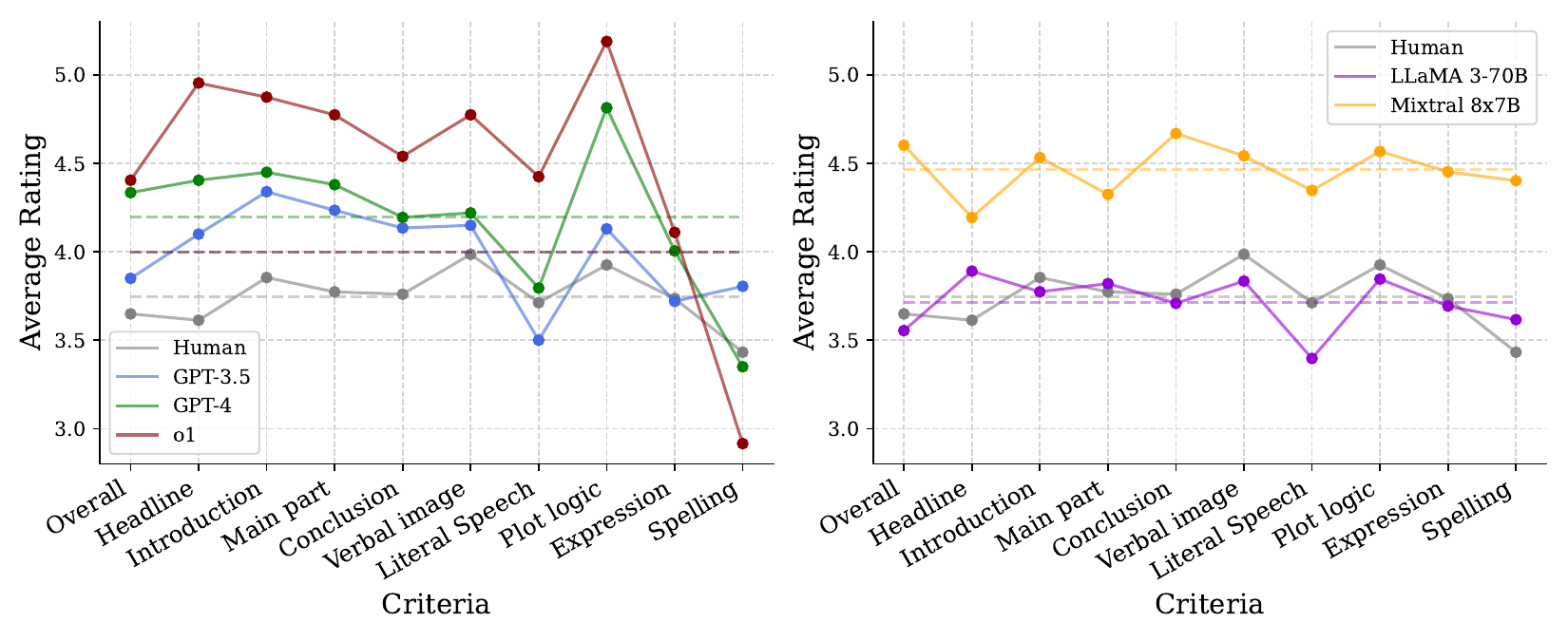}
    \caption{Average ratings for each evaluation criterion, illustrating the differences in assessments between human raters and closed-sourced LLMs (left side) and open-sources LLMs (right side).
    \label{fig:avg_rating_criteria}}
\end{figure*}

To gain further insights into the differing ratings between human raters and LLMs, we compare the average ratings for each criterion in Figure \ref{fig:avg_rating_criteria}. Generally, the GPT models exhibit higher mean ratings across all individual criteria, whereas human raters tend to be stricter in their evaluations. LLaMA 3 displays average ratings similar to those of the human raters. In contrast, the Mixtral model consistently shows significantly higher average ratings across all categories. Consequently, the differences observed in the overall ratings between LLMs and humans are reflected in each category's judgments. This consistency indicates that the final ratings align with the more detailed evaluations of human raters and LLMs.

It is important to note the variances presented in Table \ref{tab:criteria_results}. While the LLaMA 3 models exhibit average ratings comparable to human raters, their variances are considerably smaller, consistently clustering around the midpoint. Similarly, the Mixtral model also demonstrates reduced variance, aligning with the pattern observed in LLaMA 3. In contrast, GPT-3.5, o1, and human raters display higher variances, indicating a broader differentiation in their evaluations. The low ICC values for both LLaMA 3 and Mixtral models contribute to these observations, as their limited agreement among ratings results in more average values and diminished variability. 

\begin{table*}[htbp]
    \centering
    \resizebox{\textwidth}{!}{
    \begin{tabular}{lc|cc|cl|cc|cl|cc}
        \toprule
         Category & Human & GPT-3.5 & $p$ & GPT-4  &  \multicolumn{1}{c|}{$p$} & GPT-o1 & p & LLaMA 3 & \multicolumn{1}{c|}{$p$} & Mixtral & $p$ \\
         \midrule
Headline & 3.61 $\pm$ 1.16 & 4.10 $\pm$ 0.82 & 0.10  & \textbf{4.40 $\pm$ 0.60} & \textbf{0.00}  & \textbf{4.96 $\pm$ 1.12} & \textbf{0.00}  & 3.89 $\pm$ 0.45 & 0.75  & 4.19 $\pm$ 0.59 & 0.09  \\ 
Introduction & 3.85 $\pm$ 0.79 & 4.34 $\pm$ 0.91 & 0.07  & \textbf{4.45 $\pm$ 0.47} & \textbf{0.01}  & \textbf{4.87 $\pm$ 0.55} & \textbf{0.00}  & 3.77 $\pm$ 0.43 & 0.39  & \textbf{4.53 $\pm$ 0.53} & \textbf{0.00}  \\ 
Main part & 3.77 $\pm$ 0.94 & 4.24 $\pm$ 1.14 & 0.17  & \textbf{4.38 $\pm$ 0.57} & \textbf{0.01}  & \textbf{4.78 $\pm$ 0.53} & \textbf{0.00}  & 3.82 $\pm$ 0.51 & 0.53  & \textbf{4.33 $\pm$ 0.51} & \textbf{0.02}  \\ 
Conclusion & 3.76 $\pm$ 0.93 & 4.14 $\pm$ 0.91 & 0.13  & 4.20 $\pm$ 0.67 & 0.11  & \textbf{4.54 $\pm$ 0.96} & \textbf{0.00}  & 3.71 $\pm$ 0.55 & 0.45  & \textbf{4.67 $\pm$ 0.49} & \textbf{0.00}  \\ 
Verbal image & 3.99 $\pm$ 0.95 & 4.15 $\pm$ 1.08 & 0.47  & 4.22 $\pm$ 0.74 & 0.32  & \textbf{4.78 $\pm$ 0.77} & \textbf{0.01}  & 3.83 $\pm$ 0.59 & 0.63  & \textbf{4.54 $\pm$ 0.51} & \textbf{0.05}  \\ 
Literal Speech & 3.71 $\pm$ 1.07 & 3.50 $\pm$ 1.09 & 0.56  & 3.80 $\pm$ 0.48 & 0.79  & \textbf{4.42 $\pm$ 1.08} & \textbf{0.03}  & 3.40 $\pm$ 0.38 & 0.27  & \textbf{4.35 $\pm$ 0.57} & \textbf{0.02}  \\ 
Plot logic & 3.93 $\pm$ 0.88 & 4.13 $\pm$ 1.17 & 0.66  & \textbf{4.81 $\pm$ 0.59} & \textbf{0.00}  & \textbf{5.19 $\pm$ 0.68} & \textbf{0.00}  & 3.85 $\pm$ 0.51 & 0.83  & \textbf{4.57 $\pm$ 0.47} & \textbf{0.01}  \\ 
Expression & 3.74 $\pm$ 0.97 & 3.72 $\pm$ 0.98 & 0.92  & 4.01 $\pm$ 0.68 & 0.34  & 4.11 $\pm$ 0.86 & 0.24  & 3.69 $\pm$ 0.45 & 0.96  & \textbf{4.45 $\pm$ 0.50} & \textbf{0.01}  \\ 
Spelling & 3.43 $\pm$ 1.16 & 3.80 $\pm$ 1.05 & 0.36  & 3.35 $\pm$ 0.91 & 0.83  & 2.91 $\pm$ 0.86 & 0.14  & 3.62 $\pm$ 0.72 & 0.66  & \textbf{4.40 $\pm$ 0.47} & \textbf{0.00}  \\ 
\bottomrule
    \end{tabular}
    }
    \caption{\label{tab:criteria_results}Criteria-based evaluation, including $p$-values from the Mann-Whitney U test comparing the distribution of human ratings and LLM ratings for each individual category. Significant differences between human and LLM average scores are highlighted in \textbf{bold}.}
\end{table*}


To explore the mean value comparisons from Table \ref{tab:overall_results} in more detail, table \ref{tab:criteria_results} also presents the Mann-Whitney U test $p$-values, which compare the distribution of assessments across each criterion between humans and the corresponding LLM. There are no significant differences between human and GPT-4 ratings for criteria \textit{Conclusion}, \textit{Verbal Images \& Descriptive Elements}, \textit{Literal Speech \& Inner Monologue}, \textit{Expression \& Sentence Structure}, and \textit{Spelling \& Punctuation}. Except for \textit{Conclusion}, all these criteria are language-related. Similarly, o1 only exhibits a non-significant difference for \textit{Spelling \& Punctuation} and \textit{Expression}, both language-related criteria.
Since LLMs are trained on vast amounts of text data encompassing various writing styles and formalities, they are optimized for recognizing patterns and stylistic elements. Consequently, these surface-level aspects of a text, including grammar and sentence structure, can be efficiently analyzed and evaluated by LLMs in a manner similar to humans. 
Remarkably, they accurately assess learner texts despite likely limited exposure to student writings during training, demonstrating their effective generalization for educational writing evaluation.


In contrast, the discrepancy between human ratings and GPT-4 and o1 is significant for criteria \textit{Heading}, \textit{Introduction}, \textit{Main Part}, and \textit{Plot Logic}, all of which are content-related categories. This gap may arise from several factors. Although LLMs can generate coherent texts, at the moment they still lack the deep semantic understanding that humans possess. This limitation manifests in flawed logical reasoning concerning contextual details and maintaining logical consistency. Additionally, during the training process, model biases can be inherited and learned by the LLMs, resulting in milder ratings. 
Notably, their are differences between model versions, as previously indicated in Table \ref{tab:overall_results} and Figure \ref{fig:avg_rating_criteria}, which is surprising given that newer model versions typically outperforms their predecessor in complex tasks, which is expected to bring it closer to human evaluative standards. These findings highlight the ongoing challenges in aligning LLM assessments with human judgment, particularly in content-heavy criteria.


\subsection{Impact of Evaluation Categories on Overall Essay Scoring}\label{sec:intra_criteria}

\begin{figure*}[htbp]
     \centering
    \includegraphics[width=0.9\linewidth]{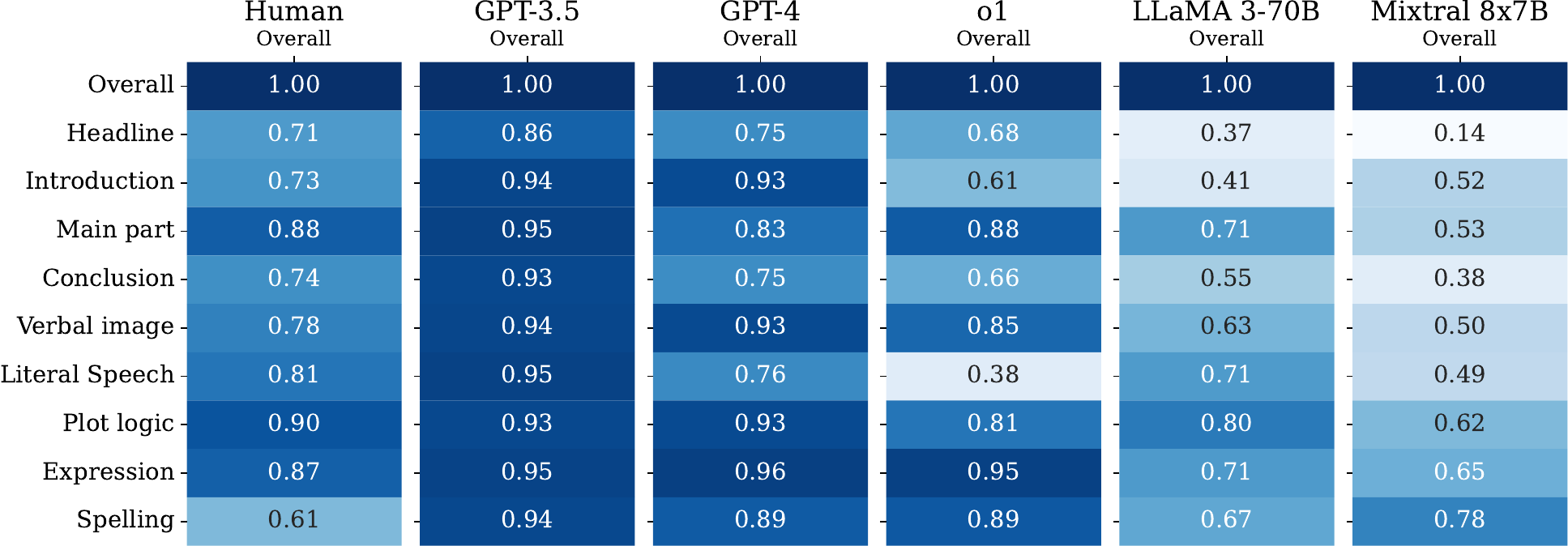}
    \caption{\label{fig:correlation matrix}Inter-criteria correlation matrix illustrating the strength of relationships between individual evaluation categories for each model. Light blue indicates weak correlations, while dark blue marks strong correlations.}
\end{figure*}


To compare how strongly individual evaluation categories impact the overall rating for teachers and LLMs (RQ3), we correlate single criteria scores with the overall rating for teachers and each model. Overall, it is noteworthy that the correlations between all features of GPT-3.5 and GPT-4 are higher than those for human judgments. This can be attributed to the auto-regressive nature of LLMs, where each new token generation is conditioned on the previous output, implicitly incorporating prior evaluations into the context. This effect is particularly pronounced in the smaller GPT-3.5 model, where correlations between ratings are nearly always \(\rho \geq 0.8\). 
The high inter-criteria correlations may indicate that the assessment of each criterion relies more on preceding ratings than solely on the original text.

In contrast, o1 exhibits fairly lower correlations between criteria, which may stem from its larger number of parameters and enhanced context understanding, allowing it to focus more on the input text than previous ratings. Compared to the previous GPT models, human raters and o1 demonstrate lower inter-criteria correlations, suggesting that their evaluations are less influenced by preceding ratings and more independently based on each criterion's merits. 

Conversely, the LLaMA 3 and Mixtral models consistently show low correlations across all criteria, highlighting a distinct evaluation pattern. This lower inter-criteria correlation suggests that the open-source model assess each criterion more independently, potentially reflecting a different internal reasoning process or a less integrated evaluation strategy compared to closed-source models.


When examining which evaluations criteria seem to highly impact overall scores, both human raters and LLMs agree that criteria \textit{Plot Logic} and \textit{Expression \& Sentence Structure} are highly relevant to the overall grade, whereas criterion \textit{Heading} is less relevant. Major differences are observed in how criterion \textit{Main Part} influences the overall grade for teachers compared to GPT-4, while o1 is able to include this information more strongly. 
Specifically, for teachers, the \textit{Main Part} is a major determinant of the overall assessment, whereas GPT-4 places greater emphasis on criterion \textit{Spelling \& Punctuation} and \textit{Verbal image}, which is not as decisive for human evaluators. This discrepancy highlights a potential risk associated with using LLMs for assessing student texts: the models may inherently focus more on linguistic surface features, such as \textit{Spelling \& punctuation}. Consequently, these surface-elements could be weighted more heavily in the overall assessment than is appropriate, potentially undermining the evaluation of deeper content quality. In addition, \textit{Main Part} and \textit{Plot Logic} are particularly influential criteria for the teachers, but precisely here Table \ref{tab:pearson_correlation_results} shows weak correlations and Table \ref{tab:criteria_results} significant differences between GPT-4, o1 and the humans. This shows one of the major challenges we are currently facing in improving machine-generated feedback.


\section{Discussion}
This study aimed to evaluate the performance and reliability of open-source and closed-source LLMs in assessing student essays, specifically focusing on German-language texts. 


\subsection{Closed-source LLMs provide reliable essay assessments}

Our analysis revealed that closed-source models, particularly o1, exhibit higher reliability when run multiple times and stronger correlations with human assessments compared to open-source models like LLaMA 3 and Mixtral. The high reliability of GPT models aligns with previous findings in the literature \cite{hackl2023gpt}. o1 demonstrated strong correlations in eight out of ten evaluation categories, aligning closely with human judgments in language-related aspects.

Conversely, open-source models like LLaMA 3 and Mixtral showed minimal to no correlation with human ratings, often rating essays around the midpoint with low variance. This limited differentiation suggests that these models are less effective in distinguishing between varying levels of essay quality, likely due to their lower ICC scores indicating poor internal consistency. This inconsistency in ratings, when run multiple times, makes them unsuitable for employment in real-world scenarios. 
\kathrin{\cite{stahl2024} find a similar lack of alignment with human evaluations with LLaMA 2 \cite{touvron2023llama}, while they were able to increase the performance of Mistral by applying prompt engineering techniques like CoT prompting \cite{wei2022chain}.}

\subsection{LLMs align best with Human Ratings in Language-Related Criteria}

The comparison of overall ratings on individual criteria revealed that GPT-4, o1, and Mixtral generally provide higher average scores than human raters, \kathrin{which is in line with previous research \cite{alers2024using,morjaria2024examining}}. The lower variance in overall scores observed in LLaMA 3 and Mixtral models indicates a tendency to rate essays uniformly, likely due to averaging highly inconsistent values of several assessments, which limits their ability to differentiate effectively between high and low-quality writing. This finding is consistent with our initial observation of low internal consistency for these models, suggesting that they lack detailed evaluation capabilities.

Moreover, the Mann-Whitney U tests indicated that the distribution of GPT-4 ratings aligns closely with human judgments in language-related criteria but diverges significantly in content-related aspects, which have the highest influence on the overall grade for real-world teachers.
This indicates a better alignment in broadly applicable linguistic features but a bigger discrepancy for content criteria, which are highly dependent on the context (i.e., grade level and text type) and on the particular standards and conventions of the German school system. This may stem from the model's lack of exposure to similar data during pre-training. Further alignment through fine-tuning or few-shot learning might mitigate this issue by providing the models with a frame of reference. Otherwise, deploying these tools in real-world educational settings could lead to inconsistent assessments and reduced usability.

\subsection{Inter-Criteria Correlations emphasize Consistent Reasoning in GPT Models}

The correlation analysis revealed that GPT models exhibit higher inter-criteria correlation among evaluation categories compared to human raters. This phenomenon can be attributed to the auto-regressive nature of LLMs, where each generated token is influenced by preceding outputs, thereby creating a more integrated and consistent reasoning process across different evaluation dimensions. Specifically, GPT-3.5 and GPT-4 showed strong correlations (\(\rho \geq 0.86\) resp. \(\rho \geq 0.75\)) between criteria, suggesting that its assessments are highly dependent on prior ratings. While this may enhance consistency, it also raises concerns about the model's ability to independently evaluate each criterion based solely on the original text.

In contrast, o1 demonstrated lower and more diverse correlations than GPT-3.5 and GPT-4, likely due to its larger parameter set and enhanced context understanding, which allow for a more balanced focus between the input text and prior evaluations. This results in a reasoning process that better reflects human evaluative practices, particularly in content-related categories. The open-source models LLaMA 3 and Mixtral, however, maintained weaker correlations across criteria, indicating a fragmented evaluation approach that lacks the cohesive reasoning seen in GPT models. These patterns suggest that the architectural and training differences between closed-source and open-source models significantly impact their evaluation strategies and reliability.

While for humans content-related criteria like main part and plot logic have the highest influence on the overall rating, GPT-4 sets a greater focus on language aspects like expression and spelling. This discrepancy points, again, to a basic limitation where LLMs may prioritize surface-level linguistic features over deeper content analysis, potentially skewing the overall assessment towards aspects like spelling and punctuation rather than substantive content quality. 

\subsection{Implications for Automated Essay Assessment}

The findings of this study underscore both the potential and the limitations of using LLMs for automated essay assessment. GPT-4 and o1, despite their high alignment with human ratings in language-related criteria, exhibit a bias towards higher overall scores. This highlights the need for careful calibration and potential alignment strategies, such as top-down prompting or bottom-up fine-tuning on domain-specific datasets, to mitigate biases and enhance the model's ability to assess essays more accurately.

Moreover, the poor performance of open-source models like LLaMA 3 and Mixtral suggests that these models are not yet ready for reliable use in educational settings without significant improvements in their consistency and alignment with human standards. The low ICC scores and minimal correlation with human ratings indicate that these models lack the necessary reliability for detailed, multidimensional assessments required in educational contexts.

\subsection{Trust and Perceived Usefulness}
\cite{ZhaiMa2022} found in their meta-study on the perceived usefulness of automated writing assessments with Chinese students that students' trust in the automated system is an important factor in the perceived usefulness of the system. Moreover, \cite{nazaretsky2022teachers} showed that it is important for teachers to understand how the decisions are made to increase trust in a system.
Additionally, \cite{Cicek2024} found that emotional trust in products could be negatively influenced by the ‘AI’ label. Against this background, it is important not to waste the potential that LLMs have for the assessment of student texts by overestimating their functionality while their development is still at the beginning.

Building on these previous findings, the high correlation and reliability of GPT-4 and o1 could foster greater trust among educators and students, enhancing the perceived usefulness of such systems. However, transparency in model limitations and ongoing efforts to align LLM assessments with human evaluative standards are essential to implementing the full potential of these technologies without limitating user confidence.

\subsection{Limitations and Future Directions}

This study has several limitations that should be acknowledged. First, while no prompt engineering was employed in this research, future work on open-source models could benefit from sophisticated strategies such as Chain-of-Thought Engineering \cite{wei2022chain}, as demonstrated in \cite{stahl2024}. \kathrin{The novel o1 series automatically incorporates CoT reasoning, which likely contributes to its superior performance.} Implementing these techniques may enhance the performance and alignment of LLMs with human ratings by enabling more detailed and context-aware evaluations. 


The scope of this study was limited to a specific set of models, namely GPT-3.5, GPT-4, o1, LLaMA 3, and Mixtral. Incorporating other models such as Claude \cite{claude} or Gemini \cite{gemini} in future research would offer a more comprehensive evaluation of LLM performance across different architectures and training paradigms. 
Furthermore, this study focused solely on one type of essay. Exploring other essay formats, such as argumentative essays, and using a more extensive and diverse dataset would enhance the generalizability and robustness of the findings. Different essay types may present unique challenges and require distinct evaluative criteria, providing a more holistic understanding of LLM capabilities in automated assessment.
Also, the inherent variance observed in multiple runs of LLMs, specifically for open-source models, suggests that real-world applications should incorporate mechanisms to aggregate multiple scores to mitigate the influence of outliers and enhance the robustness. Additionally, the absence of a gold standard due to the inherent variability among human raters poses a challenge for automated systems. 
Future studies should investigate methods to account for the fuzzy nature of human evaluations and integrate them into LLM training and assessment frameworks.

Finally, we observed a clear trend in OpenAI’s closed-source models: each new version shows improved reliability and a stronger correlation with human assessments, which was also shown by \cite{lee2024applying}. Given the rapid advancements in this field and our findings, we anticipate continued enhancements in LLMs, making them increasingly effective tools for supporting automated essay evaluation in educational settings.


\section{Conclusion}


In this study, we evaluated open-source and closed-source Large Language Models in assessing German student essays against human ratings, highlighting the strengths and limitations of multidimensional evaluation by LLMs. The novel o1 model demonstrated high reliability and strong correlations with human ratings in language-related criteria, though it tended to assign higher overall scores. In contrast, open-source models like LLaMA 3 and Mixtral showed low variance and weak correlations, limiting their effectiveness for educational assessments. These findings suggest that while o1 holds promise as an assistive tool for teachers, further refinement is necessary to enhance content evaluation and ensure alignment with human standards, especially when the goal is not just to award points on pre-determined criteria, but to provide feedback that is conducive to learning, showing students specifically what and how to improve their text.


\bibliographystyle{ACM-Reference-Format}
\bibliography{main}


\end{document}